\documentclass{article}

\usepackage[final]{nips_2018}

\usepackage[utf8]{inputenc} 
\usepackage[T1]{fontenc}    
\usepackage{hyperref}       
\usepackage{url}            
\usepackage{booktabs}       
\usepackage{amsfonts}       
\usepackage{nicefrac}       
\usepackage{microtype}      
\usepackage{algorithm}      
\usepackage{algpseudocode}  
\usepackage{graphicx}
\usepackage[inline]{enumitem}

\bibpunct{(}{)}{;}{a}{,}{,}

\AtBeginDocument{

}

\title{Reward learning from human preferences and demonstrations in Atari}

\author{
  Borja Ibarz \\
  DeepMind \\
  \texttt{bibarz@google.com} \\
  \And
  Jan Leike \\
  DeepMind \\
  \texttt{leike@google.com} \\
  \And
  Tobias Pohlen \\
  DeepMind \\
  \texttt{pohlen@google.com} \\
  \And
  Geoffrey Irving \\
  OpenAI \\
  \texttt{irving@openai.com} \\
  \And
  Shane Legg \\
  DeepMind \\
  \texttt{legg@google.com} \\
  \And
  Dario Amodei \\
  OpenAI \\
  \texttt{damodei@openai.com} \\
}
\date{\today}

\begin{document}

\maketitle

\begin{abstract}
To solve complex real-world problems with reinforcement learning, we cannot rely on manually specified reward functions.
Instead, we can have humans communicate an objective to the agent directly.
In this work, we combine two approaches to learning from human feedback: expert demonstrations and trajectory preferences.
We train a deep neural network to model the reward function and use its predicted reward to train an DQN-based deep reinforcement learning agent on 9 Atari games. Our approach beats the imitation learning baseline in 7 games
and achieves strictly superhuman performance on 2 games without using game rewards. Additionally, we investigate the goodness of fit of the reward model, present some reward hacking problems, and study the effects of noise in the human labels.
\end{abstract}

\section{Introduction}
\label{sec:introduction}

Reinforcement learning~(RL) has recently been very successful in solving hard problems in domains with well-specified reward functions~\citep{Mnih15,Mnih16,Silver16}. However, many tasks of interest involve goals that are poorly defined or hard to specify as a hard-coded reward. In those cases we can rely on demonstrations from human experts~(inverse reinforcement learning, \citealp{Ng00,Ziebart08}), policy feedback~\citep{Knox09,Warnell17}, or trajectory preferences~\citep{Wilson12,Christiano17}.

When learning from demonstrations, a policy model is trained to \emph{imitate} a human demonstrator on the task~\citep{Ho16,Hester18}. If the policy model mimics the human expert's behavior well, it can achieve the performance of the human on the task. However, to provide meaningful demonstrations, the human demonstrator has to have some familiarity with the task and understand how to perform it. In this sense, imitation learning puts more burden on the human than just providing feedback on behavior, which only requires the ability to judge outcomes.
Moreover, using this imitation learning approach it is impossible to significantly exceed human performance.

To improve on imitation learning
we can learn a reward function directly from human feedback, and optimize it using reinforcement learning.
In this work, we focus on reward learning from trajectory preferences in the same way as \citet{Christiano17}.
However, learning a reward function from trajectory preferences expressed by a human suffers from two problems:
\begin{enumerate}
\item It is hard to obtain a good state space coverage with just random exploration guided by preferences. If the state space distribution is bad, then the diversity of the trajectory that we request preferences for is low and thus the human in the loop can't convey much meaningful information to the agent.
\item Preferences are an inefficient way of soliciting information from humans, providing only a few hundred bits per hour per human.
\end{enumerate}

Our approach addresses the problems in imitation learning and learning from trajectory preferences by combining the two forms of feedback. First, we initialize the agent's policy with imitation learning from the expert demonstrations using the pretraining part of the DQfD algorithm~\citep{Hester18}. Second, using trajectory preferences and expert demonstrations, we train a reward model that lets us improve on the policy learned from imitation.

We evaluate our method on the Arcade Learning Environment~\citep{Bellemare13} because Atari games are RL problems difficult enough to benefit from nonlinear function approximation and currently among the most diverse environments for RL. Moreover, Atari games provide well-specified `true' reward functions, which allows us
to objectively evaluate the performance of our method and
to do more rapid experimentation with `synthetic' (simulated) human preferences based on the game reward.

We show that
demonstrations mitigate problem~1 by allowing a human that is familiar with the task to guide exploration consistently. This allows us to learn to play exploration-heavy Atari games such as Hero, Private Eye, and Montezuma's Revenge.
Moreover, in our experiments, using demonstrations typically halves the amount of human time required to achieve the same level of performance; demonstrations alleviate problem~2 by allowing the human to communicate more efficiently.

\subsection{Related work}
\label{subsec:related-work}

\paragraph{Learning from human feedback.}
There is a large body of work on reinforcement learning from human ratings or rankings~\citep{Wirth2017}:
\citet{Knox09}, \citet{Pilarski11}, \citet{Akrour12}, \citet{Wilson12}, \citet{Wirth13}, \citet{Daniel15}, \citet{ElAsri16}, \citet{Wirth16}, \citet{Mathewson17}, and others. Focusing specifically on deep RL, \citet{Warnell17} extend the TAMER framework to high-dimensional state spaces, using feedback to train the policy directly~(instead of the reward function). \citet{Lin17} apply deep RL from human feedback to 3D environments and improve the handling of low-quality or intermittent feedback. \citet{Saunders17} use human feedback as a blocker for unsafe actions rather than to directly learn a policy.
The direct predecessor of our work is \citet{Christiano17}, with similar tasks, rewards, policy architectures, and preference learning scheme.

\paragraph{Combining imitation learning and deep RL.}
Various work focuses on combining human demonstrations with deep RL. \citet{Hester18}, on whose method this work is based, use demonstrations to pretrain a Q-function, followed by deep Q-learning with the demonstrations as an auxiliary margin loss. \citet{Vevcerik17} apply the same technique to DDPG in robotics, and \citet{Zhang18} pretrain actor-critic architectures with demonstrations. \citet{Nair2017} combine these methods with hindsight experience replay~\citep{Andrychowicz17}. \citet{Zhu18} combine imitation learning and RL by summing an RL loss and a generative adversarial loss from imitating the demonstrator~\citep{Ho16}. Finally, the first published version of AlphaGo~\citep{Silver16} pretrains from human demonstrations. Our work differs from all these efforts in that it replaces the hand-coded RL reward function with a learned reward function; 
this allows us to employ the imitation learning/RL combination even in cases where we cannot specify a reward function.

\paragraph{Inverse reinforcement learning~(IRL).}
IRL~\citep{Ng00,abbeel2004apprenticeship,Ziebart08} use demonstrations to infer a reward function. Some versions of our method make use of the demonstrations to train the reward function---specifically, our autolabel experiments label the demonstrations as preferable to the agent policy. This is closely related to generative adversarial imitation learning~\citep{Ho16}, a form of IRL. Note, however, that in addition to training the reward function from demonstrations we also train it from direct human feedback, which allows us to surpass the performance of the demonstrator in 2 out of 9 games.

\paragraph{Reward-free learning.}
Reward-free learning attempts to avoid reward functions and instead use measures of intrinsic motivation, typically based on information theory, as a training signal~\citep{chentanez2005intrinsically,schmidhuber2006developmental,orseau2013universal}. The intrinsic motivation measure may include mutual information between actions and end states~\citep{Gregor16}, state prediction error or surprise~\citep{Pathak17}, state visit counts~\citep{storck1995reinforcement,Bellemare16}, distinguishability to a decoder~\citep{Eysenbach18}, or empowerment~\citep{salge2014empowerment}, which is also related to mutual information~\citep{mohamed2015variational}.
The present work differs from reward-free learning in that it attempts to learn complex reward functions through interaction with humans, rather than replacing reward with a fixed intrinsic objective.

\section{Method}
\label{sec:method}

\subsection{Setting}

We consider an agent that is interacting sequentially with an environment over a number of time steps~\citep{Sutton98}: in time step $t$ the agent receives an observation $o_t$ from the environment and takes an action $a_t$. We consider the episodic setting in which the agent continues to interact until a terminal time step $T$ is reached and the episode ends. Then a new episode starts. A \emph{trajectory} consists of the sequence $(o_1, a_1), \ldots (o_T, a_T)$ of observation-action pairs.

Typically in RL the agent also receives a reward $r_t \in \mathbb{R}$ at each time step. Importantly, in this work we are not assuming that such reward is available directly. Instead, we assume that there is a human in the loop who has an intention for the agent's task, and communicates this intention to the agent using two feedback channels:
\begin{enumerate}
\item \emph{Demonstrations}: several trajectories of human behavior on the task.
\item \emph{Preferences}: the human compares pairwise short trajectory segments of the agent's behavior and prefers those that are closer to the intended goal~\citep{Christiano17}.
\end{enumerate}

In our setting, the demonstrations are available from the beginning of the experiment, while the preferences are collected during the experiment while the agent is training.

The goal of the agent is to approximate as closely as possible the behavior intended by the human. It achieves this by
\begin{enumerate*}
\item imitating the behavior from the demonstrations, and
\item attempting to maximize a reward function inferred from the preferences and demonstrations.
\end{enumerate*}
This is explained in detail in the following sections.

\subsection{The training protocol}

Our method for training the agent has the following components: an \emph{expert} who provides demonstrations; an \emph{annotator}~(possibly the same as the expert) who gives preference feedback; a \emph{reward model} that estimates a reward function from the annotator's preferences and, possibly, the demonstrations; and the \emph{policy}, trained from the demonstrations and the reward provided by the reward model. The reward model and the policy are trained jointly according to the following protocol:
\begin{algorithm}[H]
\caption{Training protocol}\label{alg:preferences+demos}
\begin{algorithmic}[1]
\State The expert provides a set of demonstrations.
\State Pretrain the policy on the demonstrations using behavioral cloning using loss $J_E$.
\State Run the policy in the environment and store these `initial trajectories.'
\State Sample pairs of clips~(short trajectory segments) from the initial trajectories.
\State The annotator labels the pairs of clips, which get added to an annotation buffer.
\State (Optionally) automatically generate annotated pairs of clips from the demonstrations and add them to the annotation buffer.
\State Train the reward model from the annotation buffer.
\State Pretrain of the policy on the demonstrations, with rewards from the reward model.
\For{$M$ iterations}
    \State Train the policy in the environment for $N$ steps with reward from the reward model.
    \State Select pairs of clips from the resulting trajectories.
    \State The annotator labels the pairs of clips, which get added to the annotation buffer.
    \State Train the reward model for $k$ batches from the annotation buffer.
\EndFor
\end{algorithmic}
\end{algorithm}

Note that we pretrain the policy model twice before the main loop begins. The first pretraining is necessary to elicit preferences for the reward model. The policy is pretrained again because some components of the DQfD loss function require reward labels on the demonstrations~(see next section).

\subsection{Training the policy}

The algorithm we choose for reinforcement learning with expert demonstrations is deep Q-Learning from demonstrations (DQfD; \citealp{Hester18}), which builds upon DQN~\citep{Mnih15} and some of its extensions~\citep{Schaul15, Wang16, Hasselt16}. The agent learns an estimate of the action-value function~\citep{Sutton98} $Q(o, a)$, approximated by a deep neural network with parameters $\theta$ that outputs a set of action-values $Q(o, \cdot; \theta)$ for a given input observation $o$. This action-value function is learned from demonstrations and from agent experience, both stored in a replay buffer~\citep{Mnih15} in the form of transitions $(o_t, a_t, \gamma_{t+1}, o_{t+1})$, where $\gamma$ is the reward discount factor~(fixed value at every step except $0$ at end of an episode). Note that the transition does not include the reward, which is computed from $o_t$ by the reward model $\hat{r}$.

During the pretraining phase, the replay buffer contains only the transitions from expert demonstrations. During training, agent experience is added to the replay buffer. The buffer has a fixed maximum size, and once it is full the oldest transitions are removed in a first-in first-out manner. Expert transitions are always kept in the buffer. Transitions are sampled for learning with probability proportional to a priority, computed from their TD error at the moment they are added to and sampled from the buffer~\citep{Schaul15}.

The training objective for the agent's policy is the the cost function
$J(Q) = J_{PDDQn}(Q) + \lambda_2 J_E(Q) + \lambda_3 J_{L2}(Q)$.
The term $J_{PDDQn}$ is the prioritized~\citep{Schaul15} dueling~\citep{Wang16} double~\citep{Hasselt16} Q-loss~(PDD), combining $1$- and $3$-step returns~\citep{Hester18}. This term attempts to ensure that the $Q$ values satisfy the Bellman equation~\citep{Sutton98}. The term $J_E$ is a large-margin supervised loss, applied only to expert demonstrations. This term tries to ensure that the value of the expert actions is above the value of the non-expert actions by a given margin. Finally, the term $J_{L2}$ is an $L2$-regularization term on the network parameters. The hyperparameters $\lambda_2$ and $\lambda_3$ are scalar constants. The agent's behavior is $\epsilon$-greedy with respect to the action-value function $Q(o, \cdot; \theta)$.

\subsection{Training the reward model}

Our reward model is a convolutional neural network $\hat{r}$ taking observation $o_t$ as input~(we omit actions in our experiments)
and outputting an estimate of the corresponding reward $r_{t+1} \in \mathbb{R}$. Since we do not assume to have access to an environment reward, we resort to indirect training of this model via preferences expressed by the annotator~\citep{Christiano17}. The annotator is given a pair of \emph{clips}, which are trajectory segments of $25$ agent steps each~(approximately $1.7$~seconds long). The annotator then indicates which clip is preferred, that the two clips are equally preferred, or that the clips cannot be compared. In the latter case, the pair of clips is discarded. Otherwise the judgment is recorded in an annotation buffer $A$ as a triple $(\sigma^1, \sigma^2, \mu)$, where $\sigma^1, \sigma^2$ are the two episode segments and $\mu$ is the judgment label~(one of $(0, 1)$, $(1, 0)$ or $(0.5, 0.5)$).

To train the reward model $\hat{r}$ on preferences, we interpret the reward model as a preference predictor by assuming that the annotator's probability of preferring a segment $\sigma^i$ depends exponentially on the value of the reward summed over the length of the segment:
\[
\hat{P}[\sigma^1 \succ \sigma^2] = \exp\left(\sum\limits_{o \in \sigma^1} \hat{r}(o) \right) / \left( \exp\left(\sum\limits_{o \in \sigma^1} \hat{r}(o)\right) + \exp\left(\sum\limits_{o \in \sigma^2} \hat{r}(o) \right) \right)
\]
We train $\hat{r}$ to minimize the cross-entropy loss between these predictions and the actual judgment labels:
\[
\mathrm{loss}(\hat{r}) = -\sum_{(\sigma^1, \sigma^2, \mu) \in A} \mu(1) \log \hat{P}[\sigma^1 \succ \sigma^2] + \mu(2) \log \hat{P}[\sigma^2 \succ \sigma^1]
\]
This follows the Bradley-Terry model~\citep{Bradley52} for estimating score functions from pairwise preferences. It can be interpreted as equating rewards with a preference ranking scale analogous to the Elo ranking system developed for chess~\citep{Elo78}.

Since the training set is relatively small~(a few thousand pairs of clips) we incorporate a number of modifications to prevent overfitting: adaptive regularization, Gaussian noise on the input, L2 regularization on the output~(details in~\autoref{sec:experiment_details}). Finally, since the reward model is trained only on comparisons, its scale is arbitrary, and we normalize it every 100,000 agent steps to be zero-mean and have standard deviation $0.05$ over the annotation buffer $A$. This value for the standard deviation was chosen empirically; deep RL is very sensitive to the reward scale and this parameter is important for the stability of training.

\subsection{Selecting and annotating the video clips}
\label{sec:annotation}

The clips for annotation are chosen uniformly at random from the initial trajectories~(line 3 in \autoref{alg:preferences+demos}) and the trajectories generated during each iteration of the training protocol. Ideally we would select clips based on uncertainty estimates from the reward model; however, the ensemble-based uncertainty estimates used by \citet{Christiano17} did not improve on uniform sampling and slowed down the reward model updates. The annotated pairs are added to the annotation buffer, which stores all the pairs that have been annotated so far. The number of pairs collected after each protocol iteration decreases as the experiment progresses, according to a schedule~(see details in \autoref{sec:experiment_details}).

In some experiments we attempt to leverage the expert demonstrations to enrich the set of initial labels. In particular, each clip selected for annotation from the initial trajectories is paired with a clip selected uniformly at random from the demonstrations and a labeled pair is automatically generated in which the demonstration is preferred. Thus the initial batch of $k$ pairs of clips produces $2k$ extra annotated pairs without invoking the annotator, where $k$ is the number of labels initially requested from the annotator.

In the majority of our experiments the annotator is not a human. Instead we use a synthetic oracle whose preferences over clips reflect the true reward of the underlying Atari game. This \emph{synthetic feedback} allows us to run a large number of simulations and investigate the quality of the learned reward in some detail~(see \autoref{sec:quality_of_reward}).

\section{Experimental results}
\label{sec:results}

Our goal is to train an agent to play Atari games \emph{without access to the game's reward function}. Therefore typical approaches, such as deep RL~\citep{Mnih15,Mnih16} and deep RL with demos~\citep{Hester18} cannot be applied here. We compare the following
experimental setups~(details are provided in \autoref{sec:experiment_details}):
\begin{enumerate}
\item \emph{Imitation learning}~(first baseline). Learning purely from the demonstrations without reinforcement learning~\citep{Hester18}. In this setup, no preference feedback is provided to the agent.
\item \emph{No demos}~(second baseline). Learning from preferences without expert demonstrations, using the setup from \citet{Christiano17} with PDD~DQN instead of A3C.
\item \emph{Demos + preferences}. Learning from both preferences \emph{and} expert demonstrations.
\item \emph{Demos + preferences + autolabels}. Learning from preferences and expert demonstrations, with additional preferences automatically gathered by preferring demo clips to clips from the initial trajectories~(see \autoref{sec:annotation}).
\end{enumerate}

We've selected 9 Atari games, 6 of which~(Beamrider, Breakout, Enduro, Pong, Q*bert, and Seaquest) feature in \citet{Mnih13} and \citet{Christiano17}. Compared to previous work we exclude Space Invaders because we do not have demonstrations for it. The three additional games~(Hero, Montezuma's Revenge, and Private Eye) were chosen for their exploration difficulty: without the help of demonstrations, it is very hard to perform well in them~\citep{Hester18}.

In each experimental setup~(except for imitation learning) we compare four feedback schedules. The full schedule consists of $6800$ labels ($500$ initial and $6300$ spread along the training protocol). The other three schedules reduce the total amount of feedback by a factor of $2$, $4$ and $6$ respectively (see details in \autoref{sec:experiment_details}).

The majority of the experiments use the synthetic oracle for labeling. We also run experiments with actual human annotators in the \emph{demos + preferences} experimental setup, with the full schedule and with the schedule reduced by a factor of $2$. In our experiments the humans were contractors with no experience in RL who were instructed as in \citet{Christiano17} to only judge the outcome visible in the segments. We label these experiments as \emph{human}.

\begin{figure}[t]
\begin{center}
\includegraphics[scale=0.28]{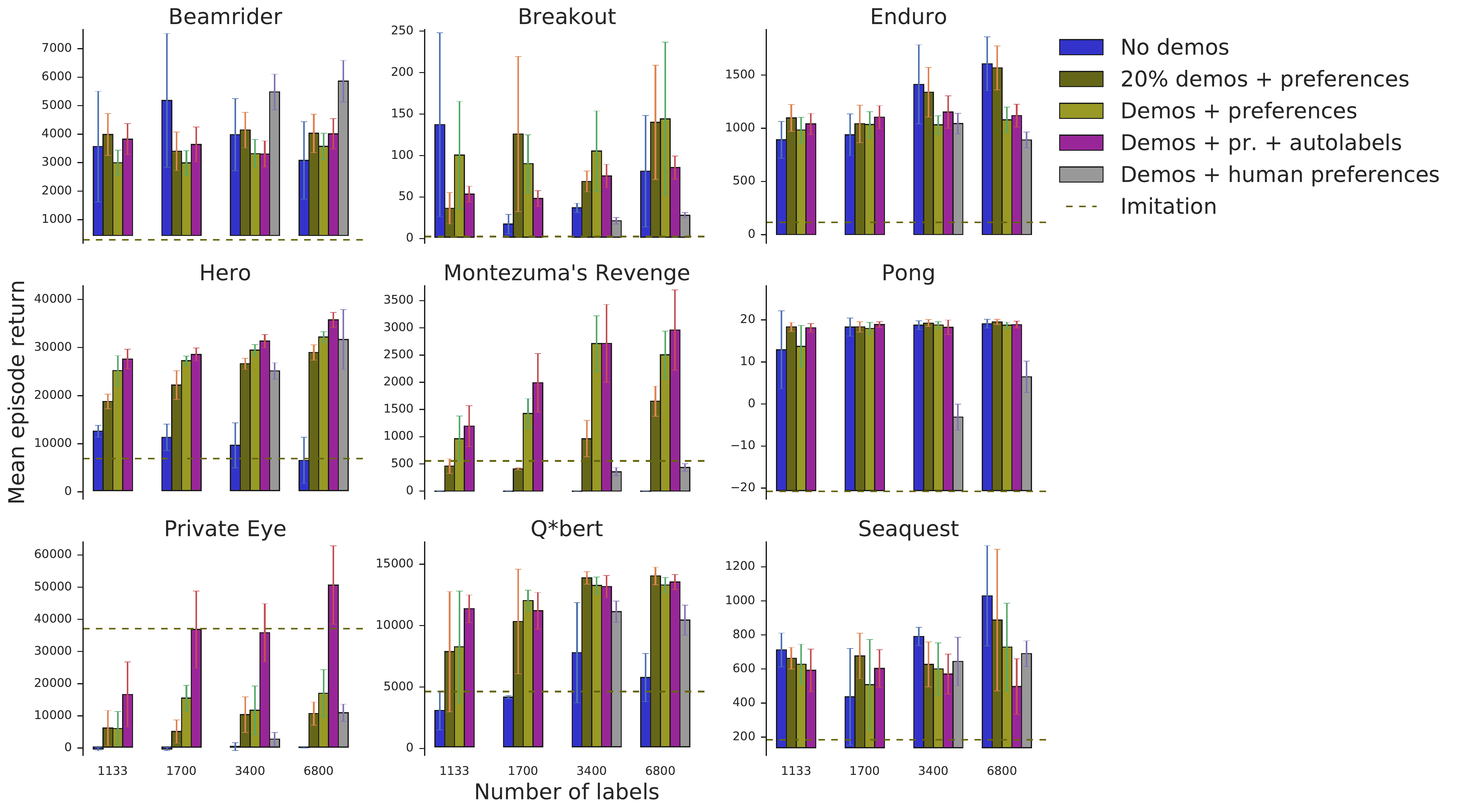}
\end{center}
\caption{Performance of our method on 9 Atari games after 50 million agent steps, for different annotation schedules and training setups: \emph{no demos} is the reward learning setup used by \citet{Christiano17}, trained with DQN; \emph{imitation} is the baseline from DQfD without RL; \emph{demos + preferences} and \emph{demos + pr. + autolables} use all demos and synthetic labels, with and without automatic labels from demos; \emph{20\% demos + preferences} is like \emph{demos + preferences} but uses only $20\%$ of the available demos; \emph{demos + human preferences} is the same setup as \emph{demos + preferences}, but with a human instead of the synthetic oracle.
The vertical lines depict the standard deviation across three runs of each experiment.
}
\label{fig:all_bars}
\end{figure}

\autoref{fig:all_bars} displays the mean episode returns in each game, setup and schedule, after $50$~million agent steps. We can compare the relative performance across four different experimental setups:

\emph{How much do preferences help~(demos + preferences vs.\ imitation)?}
  Our approach outperforms the imitation learning baseline in all games except Private Eye. In 6 of the 9 games this holds in every condition, even with the smallest amount of feedback. The bad performance of imitation learning in most Atari tasks is a known problem~\citep{Hester18} and in the absence of a reward function preference feedback offers an excellent complement. Private Eye is a stark exception: imitation is hard to beat even with access to reward~\citep{Hester18}, and in our setting preference feedback is seriously damaging, except when the demonstrations themselves are leveraged for labeling.

\emph{How much do demos help~(demos + preferences vs.\ no demos)?}
  Hero, Montezuma's Revenge, Private Eye and Q*bert benefit greatly from demonstrations. Specifically, in Montezuma's Revenge and Private Eye there is no progress solely from preference feedback; without demonstrations Hero does not benefit from increased feedback; and in Q*bert demonstrations allow the agent to achieve better performance with the shortest label schedule~($1100$ labels) than with the full no-demos schedule. With just $20\%$ of the demonstrations~(typically a single episode) performance already improves significantly\footnote{Experiments with $50\%$ of the demonstrations~(not shown) produced scores similar to the full demo experiments---the benefits of demonstration feedback seem to saturate quickly.}. In the rest of the games the contribution of demonstrations is not significant, except for Enduro, where it is harmful, and possibly Seaquest. In Enduro this can be explained by the relatively poor performance of the expert: this is the only game where the trained agents are superhuman in all conditions.
  Note that our results for \emph{no demos} are significantly different from those in \citet{Christiano17} because we use DQN~\citep{Mnih15} instead of A3C~\citep{Mnih16} to optimize the policy~(see \autoref{sec:compare_a3c}).

\emph{How does human feedback differ from the synthetic oracle~(demos + preferences vs.\ human)?}
  Only in Beamrider is human feedback superior to synthetic feedback~(probably because of implicit reward shaping by the human). In most games performance is similar, but in Breakout, Montezuma's Revenge and Pong it is clearly inferior. This is due to attempts at reward shaping that produce misaligned reward models~(see \autoref{fig:reward_alignment} and \autoref{sec:reward_alignment}) and, in the case of Montezuma's Revenge, to the high sensitivity of this game to errors in labeling~(see \autoref{sec:label_noise}).

\emph{How much do automatic preference labels help~(demos + preference vs.\ demos + preferences + auto labels)?}
  Preference labels generated automatically from demonstrations increase performance in Private Eye, Hero, and Montezuma's Revenge, where exploration is difficult. On most games, there are no significant differences, except in Breakout where human demonstrations are low quality~(they do not `tunnel behind the wall') and thus hurt performance.

\subsection{Use of human time}
\label{sec:human_time}

\begin{figure}[t]
\begin{center}
\includegraphics[scale=0.25]{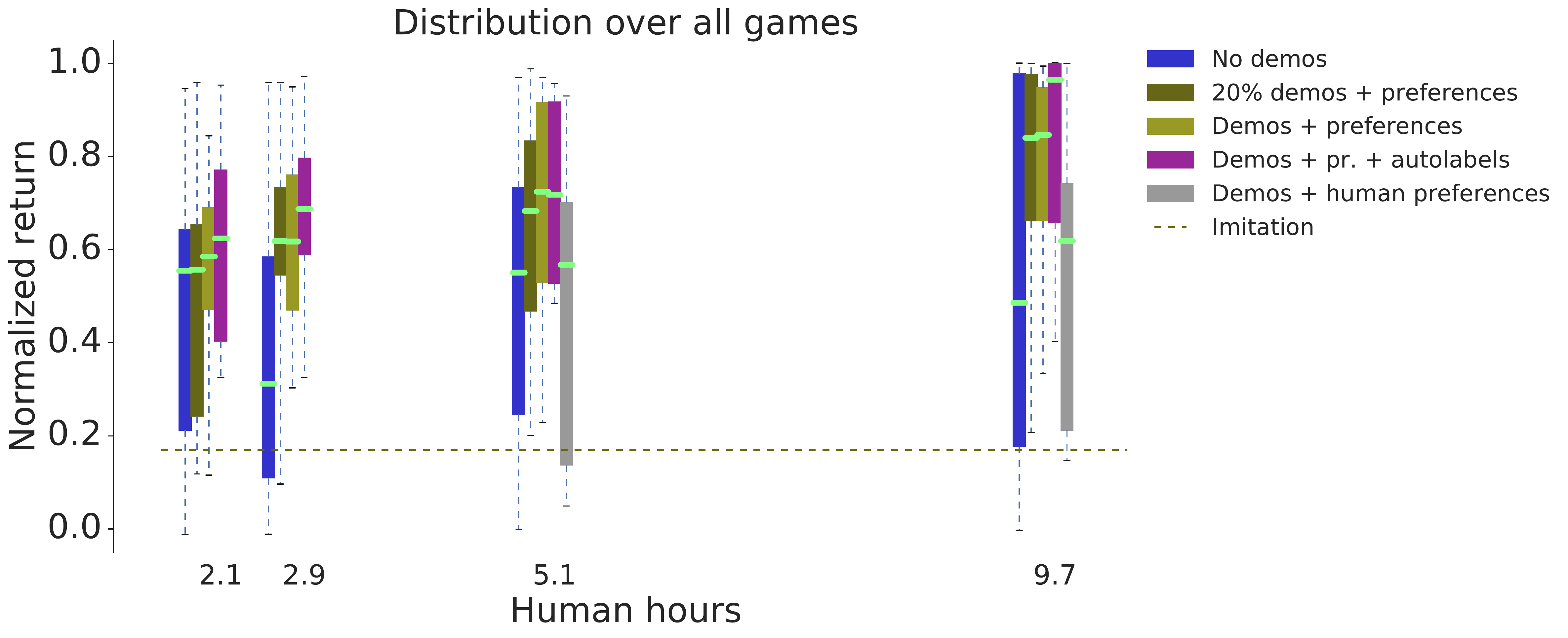}
\end{center}
\caption{Aggregated performance over all games after 50 million agent steps for different schedules and training setups. Performance is normalized for each game between 0 (return of a random policy) and 1 (best return across all setups and schedules). The boxplots show the distribution over all 9 games, the bright notch representing the median, boxes reaching the $25$ and $75$ percentiles, and whiskers the whole range. Their position along the $x$ axis shows with the total number of annotation labels used.
}
\label{fig:boxplots}
\end{figure}

\autoref{fig:boxplots} summarizes the overall performance of each setup by human time invested. More than half of the games achieve the best performance with full feedback and the help of demonstrations for imitation and annotation, and, for each feedback schedule, the majority of games benefit from demonstrations, and from the use of demonstrations in annotation. With only $3400$ labels even the worst-performing game with demonstrations and automatic labels beats the median performance without demonstrations and the full $6800$ labels. If demonstrations are not available there are games that never go beyond random-agent scores; demonstrations ensure a minimum of performance in any game, as long as they are aided by some preference feedback.
For further details refer to \autoref{app:human-effort}.

\subsection{Quality of reward model}
\label{sec:quality_of_reward}

In our experiments we are evaluating the agent on the Atari game score, which may or may not align with the reward from the reward model that the agent is trained on. With synthetic labels the learned reward should be a good surrogate of the true reward, and bad performance can stem from two causes:
\begin{enumerate*}[label={(\arabic*)}]
\item failure of the reward model to fit the data, or
\item failure of the agent to maximize the learned reward.
With human labels there are two additional sources of error:
\item mislabeling and
\item a misalignment between the true~(Atari) reward function and the human's reward function.
\end{enumerate*}
In this section we disentangle these possibilities.

\begin{figure}[t]
\begin{center}
\includegraphics[scale=0.28]{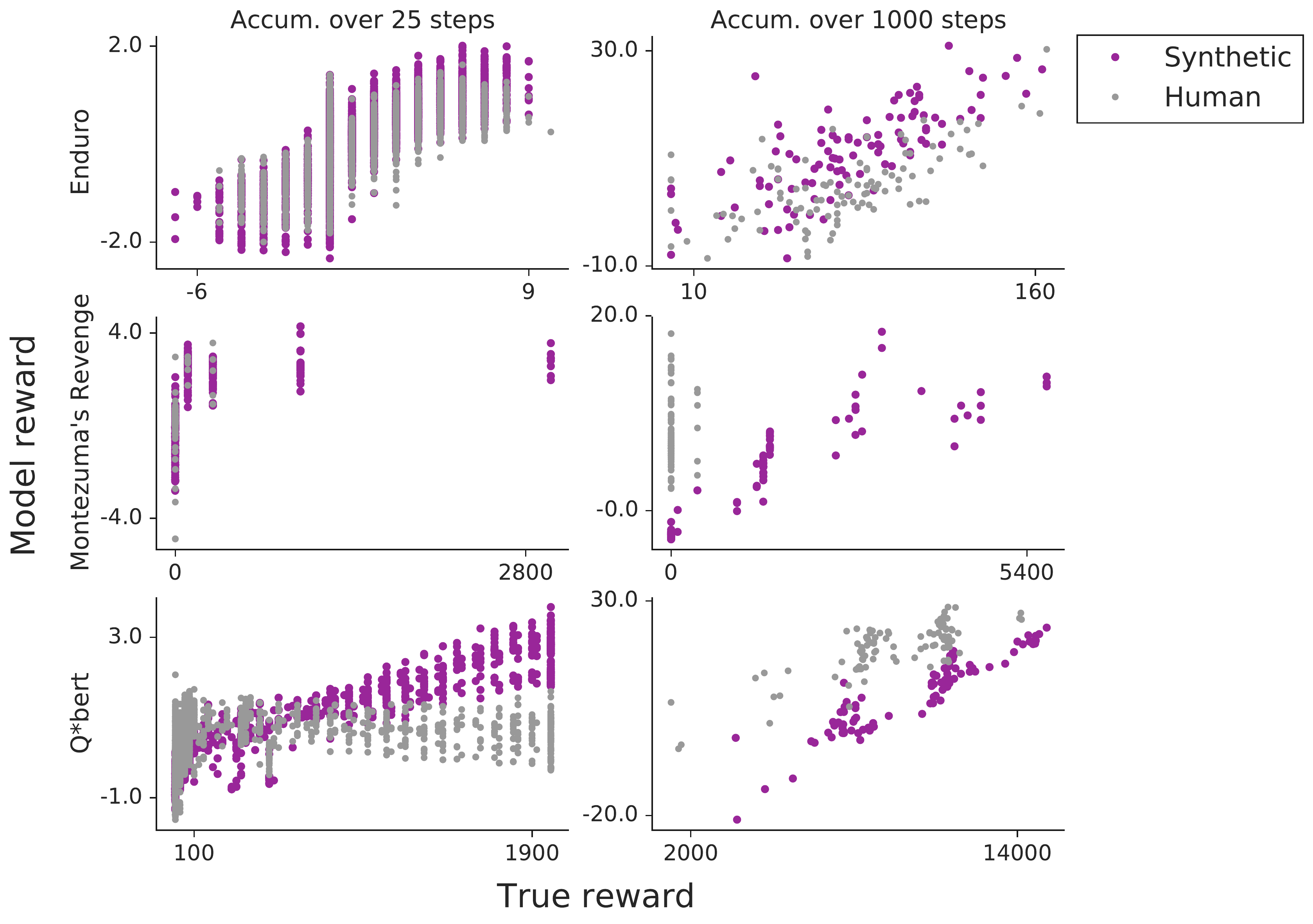}
\end{center}
\caption{True vs.\ learned reward accumulated in sequences of $25$~(left) and $1000$~(right) agent steps in Enduro, Montezuma's Revenge and Q*bert. Magenta and gray dots represent the model learned from synthetic and human preferences, respectively. A fully aligned reward model would have all points on a straight line. For this evaluation, the agent policy and reward model were fixed after successful full-schedule training~(in the case of synthetic preference feedback we chose the most successful seed; in the case of human preference feedback only one run was available).
}
\label{fig:reward_alignment}
\end{figure}

Learning the reward model is a supervised learning task, and in \autoref{sec:predictor_loss} we argue that it succeeds in fitting the data well. \autoref{fig:reward_alignment} compares the learned reward model with the true reward in three games~(see \autoref{sec:reward_alignment} for the other six games). Both synthetic~(\emph{demos + pr.\ + autolabels} in \autoref{fig:all_bars}) and human preference models are presented for comparison. Perfect alignment between true reward and modelled reward is achieved if they are equal up to an affine-linear transformation; in this case all points in the plot would be on a straight line. In most games the synthetically trained reward model is reasonably well-aligned, so we can rule out cause (1).

In Enduro both human and synthetic preferences produce well-aligned reward models, especially over long time horizons.
Q*bert presents an interesting difference between human and synthetic preferences: on short timescales, the human feedback does not capture fine-grained reward distinctions~(e.g., whether the agent covered one or two tiles) which are captured by the synthetic feedback. However, on long timescales this does not matter much and both models align well. A similar pattern occurs in Hero. Finally, in Montezuma's Revenge human feedback fails while synthetic feedback succeeds. This is partially due to a misalignment (because the human penalizes death while the Atari score does not) and partially due to the sensitivity of the reward function to label noise.
For more details, see \autoref{sec:reward_alignment}.

The difference between synthetically and human-trained reward model captures causes (3) and (4). To disentangle (3) and (4), we also provide experiments with a mislabeling rate in \autoref{sec:label_noise}.

\begin{figure}[ht]
\begin{center}
\includegraphics[scale=0.22]{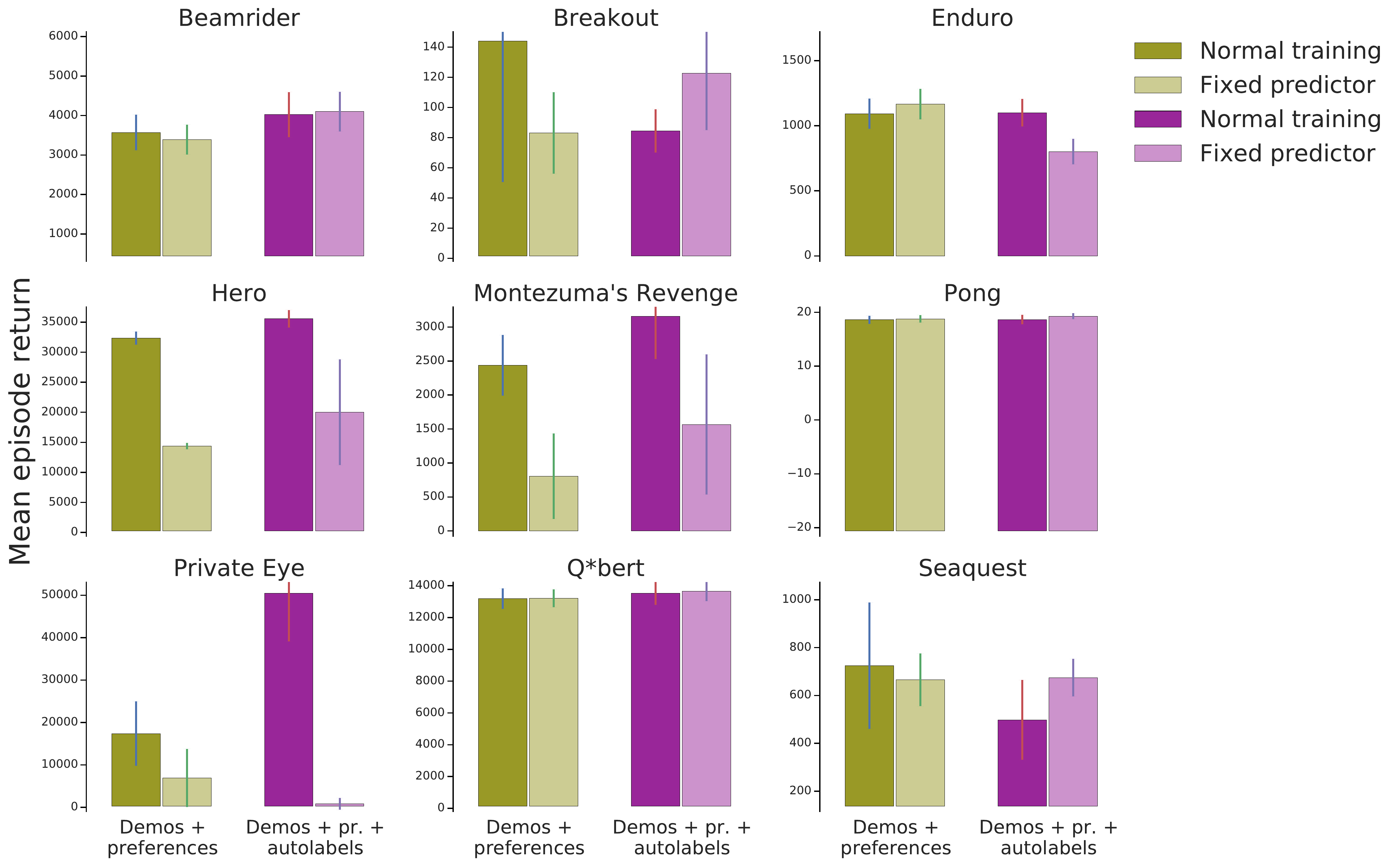}~~~~~
\includegraphics[scale=0.22]{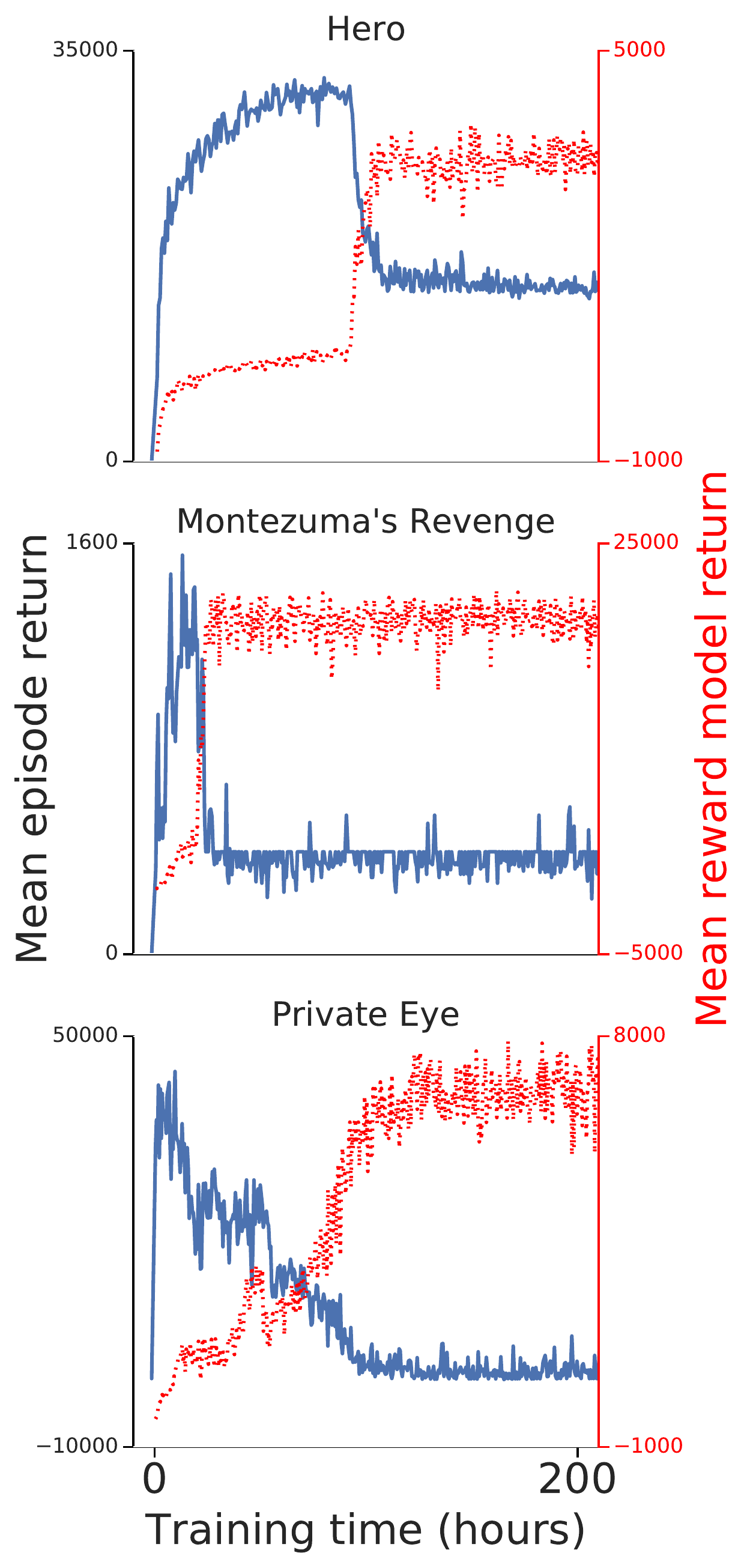}
\end{center}
\caption{Failure modes when training from a frozen reward model~(contrary to our method).
Left: performance at each game after 50 million agent steps. The darker colored bars show the results from our training protocol~(same as \autoref{fig:all_bars}) with the full label schedule. The reward model from the best seed in these experiments is then frozen and used to train an agent from scratch, resulting in the lighter colored bars.
Right: average true return~(blue) and average reward model return~(red) during training of three games~(only one seed shown per game) from a frozen reward model. This showcases how the agent learns to exploit the reward model: over time the perceived performance~(according to the reward model) increases, while the actual performance~(according to the game score) plummets.
}
\label{fig:fixed_predictor}
\vspace{-3.5mm}  
\end{figure}

\paragraph{Reward hacking.}
To further evaluate the quality of the reward model, we run experiments with frozen reward models obtained from successful runs. The result is shown in \autoref{fig:fixed_predictor}, left. Although a fully trained model should make learning the task easier, in no case is the fixed-model performance significantly better than the online training performance, which suggests that joint training of agent and reward is not intrinsically problematic. Moreover, in Hero, Montezuma, and Private Eye the performance with a fully trained reward model is much worse than online reward model training. In these cases the drop in performance happens when the agent learns to exploit undesired loopholes in the reward function~(\autoref{fig:fixed_predictor}, right), dramatically increasing the predicted reward with behaviors that diminish the true score.\footnote{Videos at \url{https://youtube.com/playlist?list=PLehfUY5AEKX-g-QNM7FsxRHgiTOCl-1hv}} These loopholes can be fixed interactively when the model is trained online with the agent, since exploitative behaviors that do not lead to good scores can be annotated as soon as they feature significantly in the agent's policy, similar to adversarial training~\citep{Goodfellow14}. With online training we also observed cases where performance temporarily drops, with simultaneous increases in model reward, especially when labels are noisy~(\autoref{sec:label_noise}).

\section{Discussion}
\label{sec:discussion}

Combining both preferences and demonstrations outperforms using either in isolation.
Their combination is an effective way to provide guidance to an agent in the absence of explicit reward~(\autoref{fig:all_bars}). Even small amounts of preference feedback~(about $1000$ comparisons) let us outperform imitation learning in 7 out of 9 games. Moreover, the addition of demonstrations to learning from preferences typically results in substantial performance gains, especially in exploration-heavy games. We achieve superhuman performance on Pong and Enduro, which is impossible even with perfect imitation.

Synthetic preference feedback proved more effective than feedback provided by humans. It could be expected that human feedback has the advantage in the exploration-heavy games, where the human can shape the reward to encourage promising exploration strategies. Analysis of the labels shows that the human annotator prefers clips where the agent seems to be exploring in particular directions. However,
instead of encouraging exploration, this feedback produces `reward pits' that trap the agent into repetitive and fruitless behaviors. This effect is not novel; \citet{MacGlashan17} have previously argued that humans are bad at shaping reward. However, our results show that demonstrations can provide consistent exploration guidance.

In addition to the experiments presented here, we were unsuccessful at achieving significant performance improvements from a variety of other ideas: distributional RL~\citep{Bellemare17}, quantile distributional RL~\citep{Dabney17}, weight sharing between reward model and policy, supplying the actions as input to the reward model, pretrained convolutional layers or semi-supervised training of the reward model, phasing out of the large-margin supervised loss along training, and other strategies of annotation from demos~(see \autoref{sec:unsuccessful-ideas}).

In contrast to \citet{Christiano17}, whose work we build upon here, we use the value-based agent DQN/DQfD instead of the policy-gradient-based agent A3C.
This shows that learning reward functions is feasible across two very different RL algorithms with comparable success. \autoref{sec:compare_a3c} compares the scores of the two agents.

Finally, \autoref{sec:quality_of_reward} highlights a caveat of reward learning: sometimes the agent learns to exploit unexpected sources of reward. This so-called \emph{reward hacking} problem~\citep{Amodei16,everitt2018} is not unique to reward learning; hard-coded reward functions are also exploitable in this way~\citep{Lehman18}. Importantly, we only found persistent reward hacking when the preference feedback was frozen. This suggests that our method, keeping a human in the training loop who provides \emph{online} feedback to the agent, is effective in preventing reward hacking in Atari games.

\subsection*{Acknowledgements}

We thank Serkan Cabi, Bilal Piot, Olivier Pietquin, Tom Everitt, and Miljan Martic for helpful feedback and discussions. Moreover, we thank Elizabeth Barnes for proofreading the paper and Ashwin Kakarla, Ethel Morgan, and Yannis Assael for helping us set up the human experiments. Last but not least, we are grateful to the feedback annotators for their many hours of meticulous work.

\bibliography{references}

\newpage\appendix
\section{Experimental details}
\label{sec:experiment_details}

\subsection{Environment}
We use the Arcade Learning Environment~\citep{Bellemare13} with the standard set of environment wrappers used by \citet{Mnih15}: full set of $18$ actions, $0$ to $90$ no-ops in the beginning of an episode, max-pooling over adjacent frames with action repeat and frame stacking of $4$ frames, observation resizing to $84$x$84$ and converting to grayscale. Across this paper we treat $4$ frames as one observation with action repeat as a single actor step, i.e., one actor step corresponds to $4$ Atari frames.

We replace the score with a constant black background to prevent inferring the reward from the score. Life loss and end-of-episode signals are not passed to the agent, effectively converting the environment into a single continuous episode. When providing synthetic oracle feedback we replace episode ends with a penalty in all games except Pong; the agent must learn this penalty.

\subsection{Expert demonstrations}
\label{sec:demos}
The same set of demonstrations for each game is used in all experiments. Our demonstrations are the same demonstrations as used by \citet{Hester18} and were collected from an expert game tester. The length and scores of these demonstrations are as follows.

\begin{center}
\begin{tabular}{p{0.23\textwidth}|p{0.1\textwidth}|p{0.1\textwidth}|p{0.1\textwidth}|p{0.1\textwidth}|p{0.11\textwidth}}
Game & Episodes & Transitions & Avg score & Min score & Max score \\
\hline
Beamrider & 4 & 38665 & 16204 & 12594 & 19844 \\
Breakout & 9 & 10475 & 38 & 17 & 79 \\
Enduro & 5 & 42058 & 641 & 383 & 803 \\
Hero & 5 & 32907 & 71023 & 35155 & 99320 \\
Montezuma's Revenge & 5 & 17949 & 33220 & 32300 & 34900 \\
Pong & 3 & 17719 & -8 & -12 & 0 \\
Private Eye & 5 & 10899 & 72415 & 70375 & 74456 \\
Q*bert & 5 & 75472 & 89210 & 80700 & 99450 \\
Seaquest & 7 & 57453 & 74374 & 56510 & 101120 \\
\end{tabular}
\end{center}

\subsection{Agent and reward model}

We optimize policies using the DQfD algorithm~\citep{Hester18}, with standard architecture and parameters: dueling, double Q-learning with a target network updated every $8000$ actor steps, discount $\gamma=0.99$, a mix of $1$- and $3$-step returns, prioritized replay~\citep{Schaul15} based on TD error with exponent $\alpha=0.5$ and importance sampling exponent $\beta=0.4$. The buffer size of $1e6$ for actor experience plus permanent demonstrations, batch size $32$, learning update every $4$ steps, additional large-margin supervised loss for expert demonstrations (Q margin=$1$, loss weight=$1$), priority bonus for expert demonstrations $\epsilon_d=3$. We stack 4 steps as input to the Q-value network. The optimizer is Adam~\citep{Kingma14} with learning rate $0.0000625$, $\beta_1=0.9$, $\beta_2=0.999$, $\epsilon=0.00015625$. Importantly, each time a batch is sampled from the buffer for learning, the reward values corresponding to the batch are computed using the reward model.

The DQfD agent policy is $\epsilon$-greedy with epsilon annealed linearly from $0.1$ to $0.01$ during the
first $10^5$ actor steps.

For the reward model, we use the same configuration as the Atari experiments in \citet{Christiano17}: $84$x$84$x$4$ stacked frames (same as the inputs to the policy) as inputs to $4$ convolutional layers of size $7$x$7$, $5$x$5$, $3$x$3$, and $3$x$3$ with strides $3$, $2$, $1$, $1$, each having $16$ filters, with leaky ReLU nonlinearities ($\alpha=0.01$). This is followed by a fully connected layer of size $64$ and then a scalar output. The agent action $a_t$ is not used as input as this did not improve performance. Since the training set is relatively small~(a few thousand pairs of clips) we incorporate a number of modifications to this basic approach to prevent overfitting:

\begin{itemize}
\item A fraction of $1/e$ of the data is held out to be used as a validation set. We use L2-regularization of network weights with the adaptive scheme described in \citet{Christiano17}: the L2-regularization weight increases if the average validation loss is more than $50\%$ higher than the average training loss, and decreases if it is less than $10\%$ higher (initial weight $0.0001$, multiplicative rate of change $0.001$ per learning step).
\item An extra loss proportional to the square of the predicted rewards is added to impose a zero-mean Gaussian prior on the reward distribution.
\item Gaussian noise of amplitude $0.1$~(the grayscale range is $0$ to $1$) is added to the inputs.
\item Convolutional layers use batch normalization~\citep{Ioffe15} with decay rate $0.99$ and per-channel dropout~\citep{Srivastava14} with $\alpha=0.8$.
\item We assume there is a $10$\% chance that the annotator responds uniformly at random, so that the model will not overfit to possibly erroneous preferences. We account for this error rate by using $\hat{P}_e = 0.9 \hat{P} + 0.05$ instead of $\hat{P}$ for the cross-entropy computation.
\end{itemize}
Finally, since the reward model is trained merely on comparisons, its absolute scale is arbitrary. Therefore we normalize its output so that it has $0$ mean and standard deviation $0.05$ across the annotation buffer. We do \emph{not} use an ensemble of reward models as done by \citet{Christiano17}. The model is trained on batches of $16$ segment pairs~(see below), optimized with Adam~\citep{Kingma14} with learning rate $0.0003$, $\beta_1=0.9$, $\beta_2=0.999$, and $\epsilon=10^{-8}$.

The training set for the reward model~(the annotation buffer) consists of annotated pairs of clips, each of $25$ actor steps~($1.7$ seconds long at $15$ fps with frame skipping). The size of the training set grows over the course of the experiment as segments are collected according to an annotation schedule. The `full' annotation schedule consists of a set of $500$ labels from initial trajectories at the beginning of training, followed by a decreasing rate of annotation, roughly proportional to $5 \cdot 10^6/(T + 5\cdot 10^6)$, where $T$ is the number of actor steps. The total number of labels in a $50$ million-step experiment is $6800$. We compare performance with proportionally reduced schedules that have $2$, $4$ and $6$ times fewer labels than the full schedule~(including the initial batch), with a total number of labels of $3400$, $1700$ and $1133$ respectively.

\subsection{Training protocol}
The training protocol consists of $500$ iterations and each iteration consists of $10^5$ agent steps. The reward model is fixed during each iteration. Trajectories~(effectively one long episode per iteration, since we removed episode boundaries) are collected in each iteration and clips of $25$ agent steps are picked at random for annotation. When using synthetic annotation, clip pairs are labeled and added to the annotation set immediately after each iteration. In experiments with preference feedback from real human annotators, clips are labeled about every $6$ hours, corresponding to about $12$ iterations. After each iteration, the reward model is trained for $6250$ batches sampled uniformly from the annotation buffer.

A pretraining phase precedes the training iterations. The pretraining phase consists of the following:

\begin{itemize}
\item Training the agent purely from demonstrations. This includes both the imitation large-margin loss and the Q loss from expert demonstrations. Notice that, since the reward model has not yet been trained at this point, it predicts small random values, and the Q loss, which is based on those predicted rewards, is noisy and acts as a regularizer. In this phase of pretraining we train on $20,000$ batches.
\item Generating trajectories from the policy trained in the previous steps to collect the initial set of $500$ clip pairs~($250$, $125$ or $83$ in the case of reduced schedules).
\item Labeling the initial set of clip pairs. In some experiments, an additional set of $1000$ ($500$, $250$, $167$ for reduced schedules) labeled pairs is automatically generated by comparing each clip in each pair of the initial set with a clip sampled uniformly at random from the demonstrations. It is automatically labeled to prefer the clip from the demonstration.
\item Training the reward model with $50000$ batches from the labeled clips.
\item Another round of agent training purely from demonstrations. Unlike the first pretraining phase, the reward model has now undergone some training and the Q loss is more meaningful. This last phase of pretraining consists of $60000$ batches.
\end{itemize}

\section{Performance as a function of human effort}
\label{app:human-effort}

\begin{figure*}[h!]
\begin{center}
\includegraphics[scale=0.18]{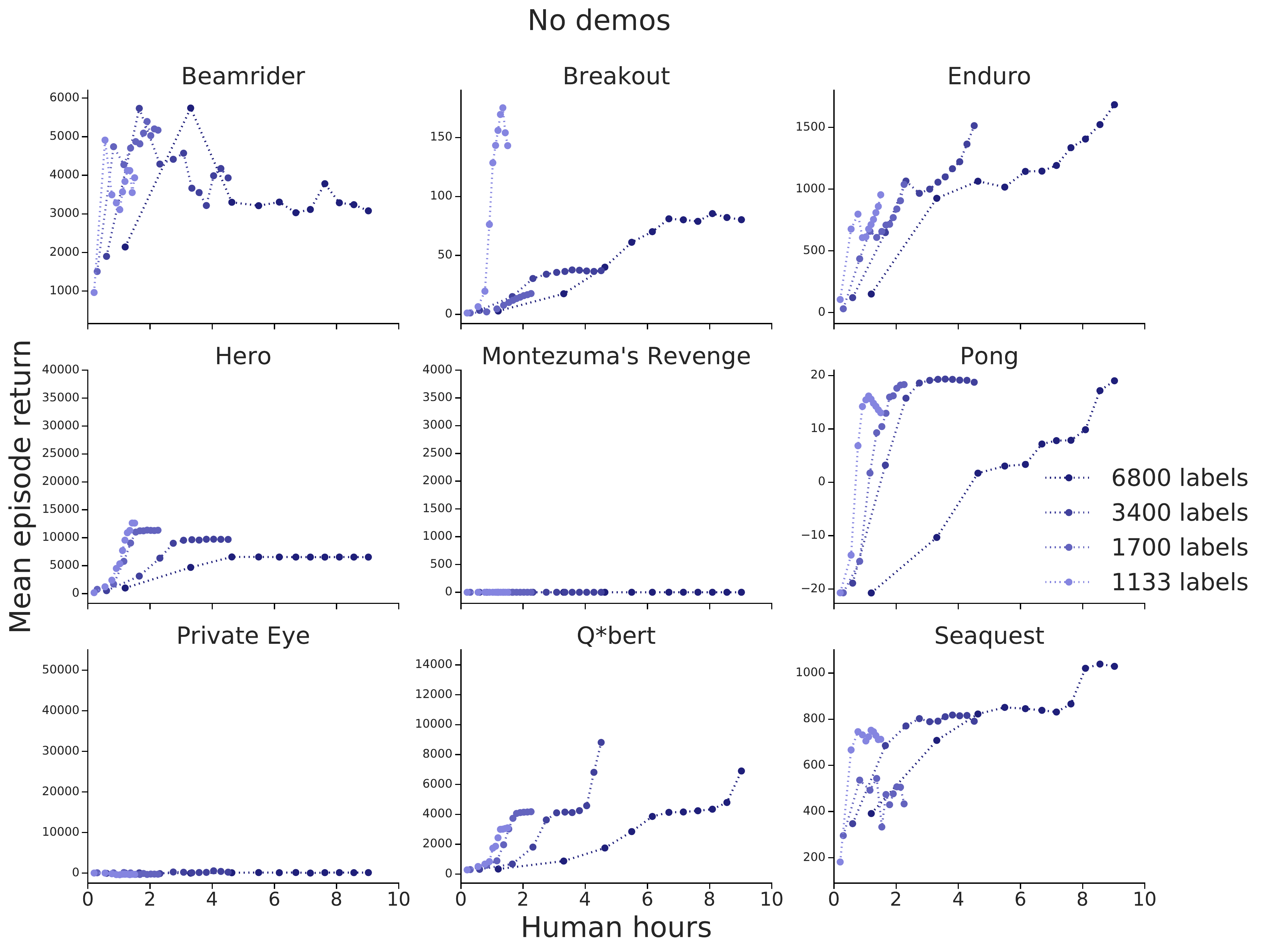}
\includegraphics[scale=0.18]{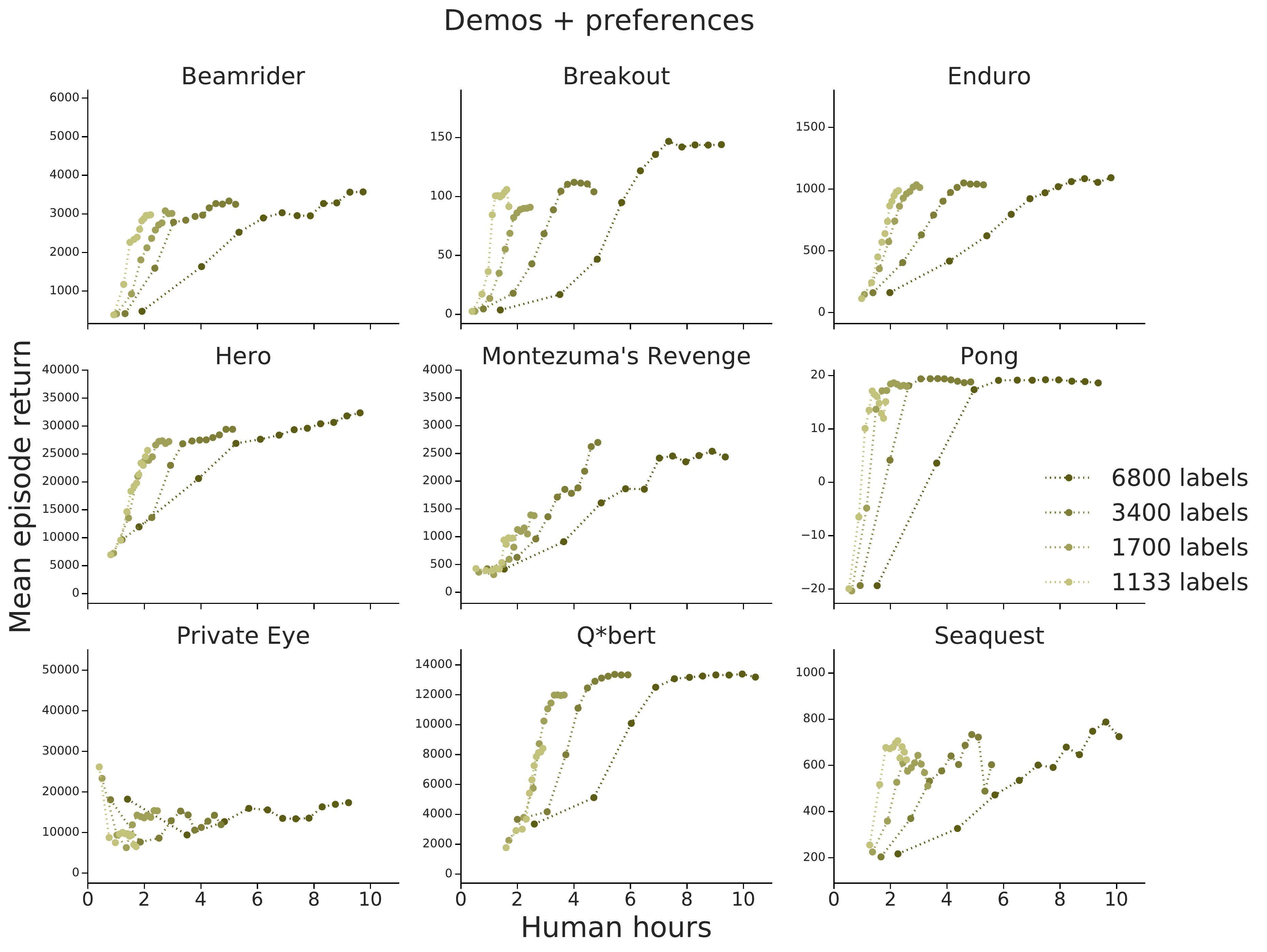}
\includegraphics[scale=0.18]{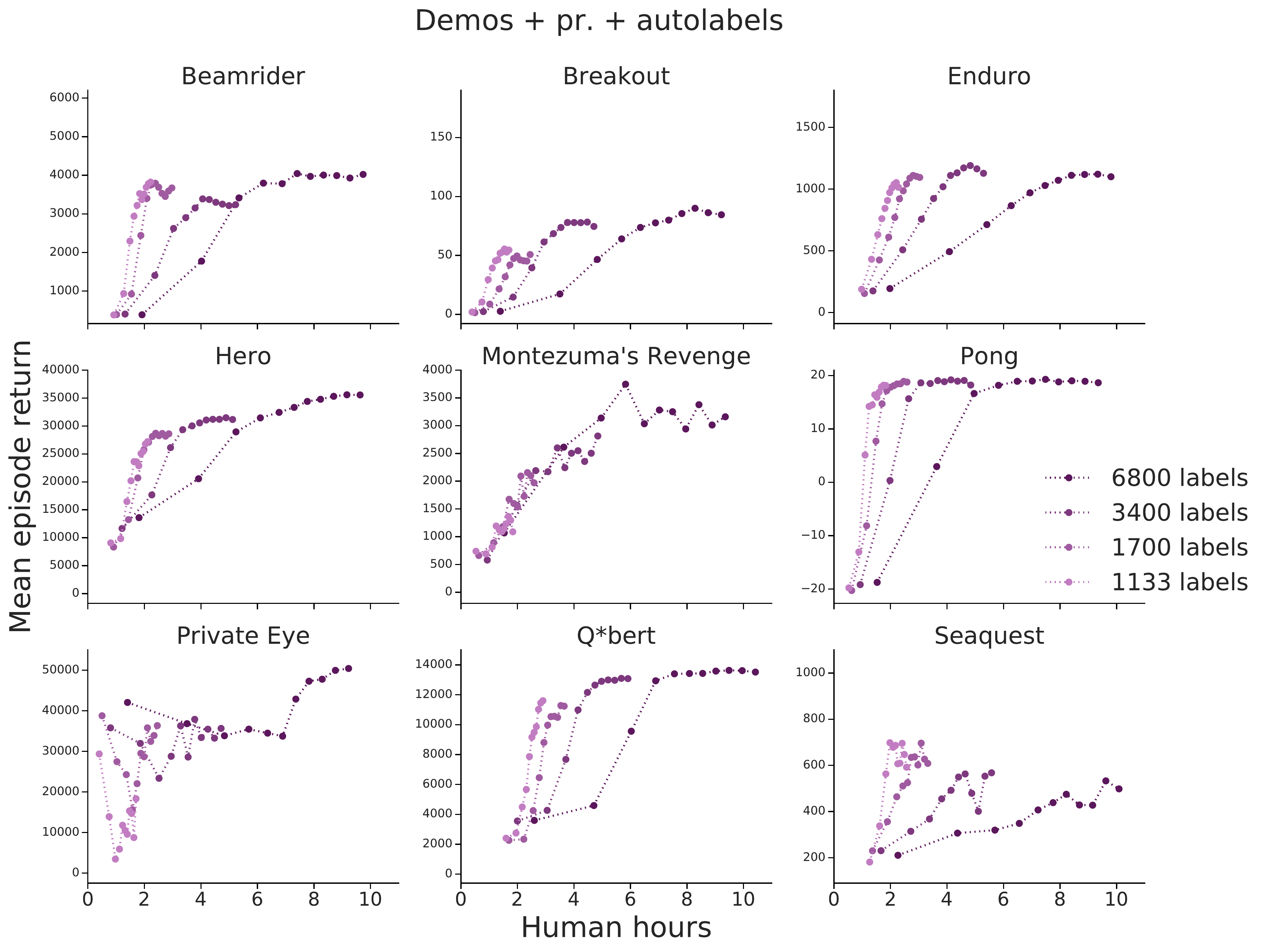}
\includegraphics[scale=0.18]{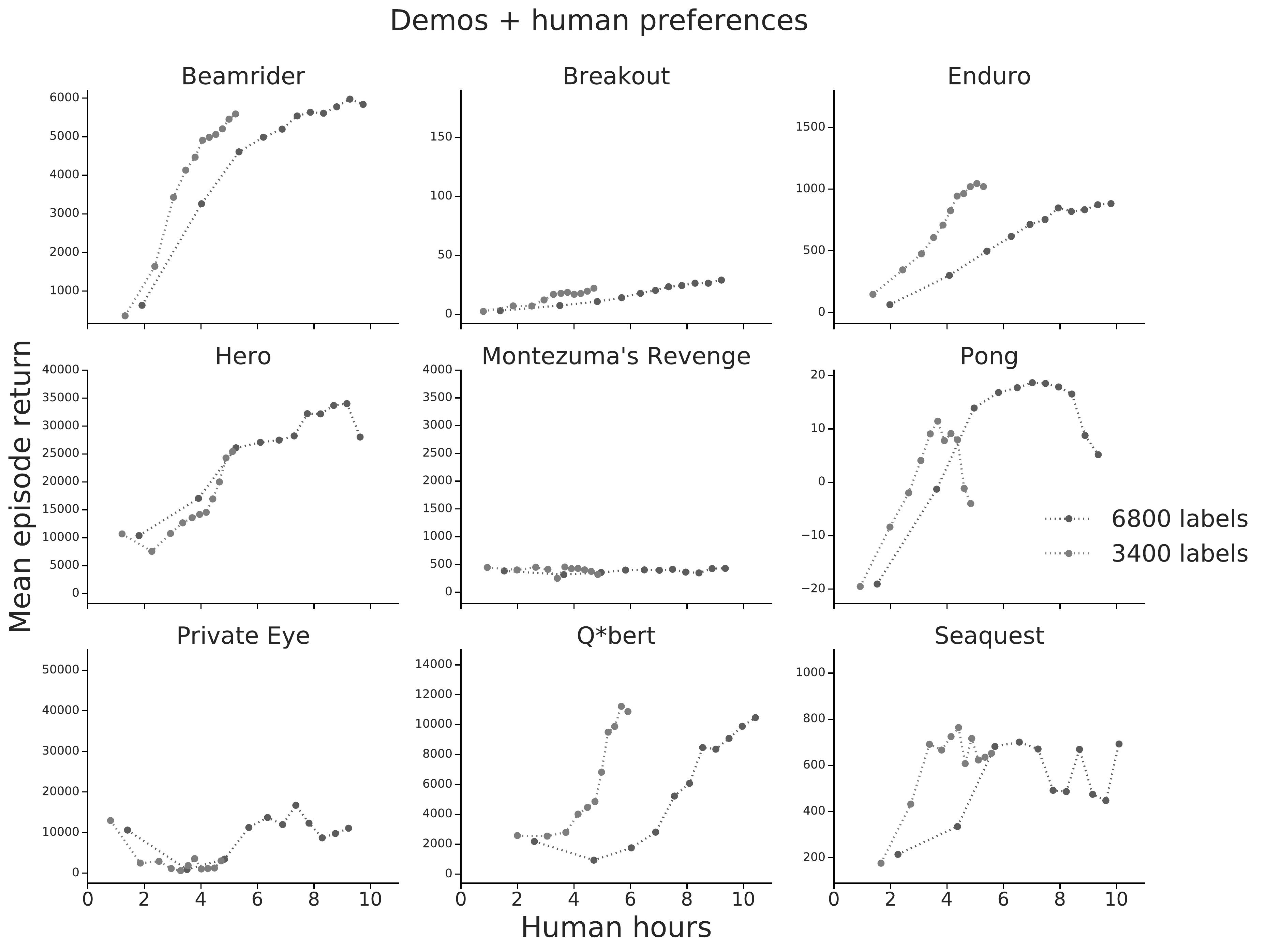}
\end{center}
\caption{Performance at each game as a function of human~(or synthetic) effort, adding labeling time~(at 750 labels/hour) and demonstration time~(at 15 fps).}
\label{fig:effort_detail}
\end{figure*}

\begin{figure*}[h!]
\begin{center}
\includegraphics[scale=0.25]{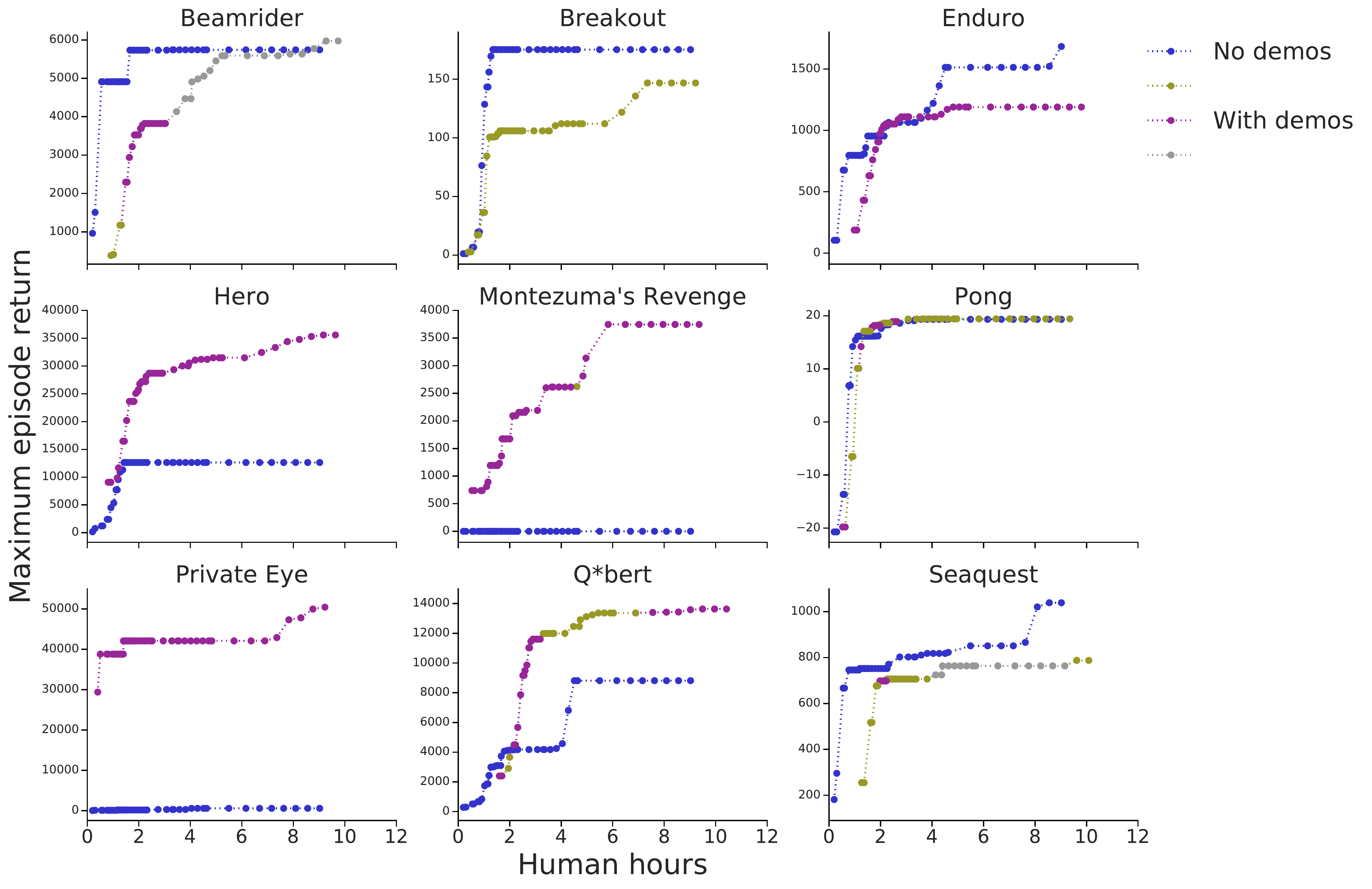}
\end{center}
\caption{Best performance up to a given human effort (in terms of time) for each game, with and without demonstrations. The different colors of the with-demos lines correspond to the best setup, as displayed in \autoref{fig:effort_detail}}
\label{fig:effort_time_max}
\end{figure*}

\autoref{fig:effort_detail} shows the performance in each game as a function of joint labeling and demonstration effort~(measured in human hours), for the different preference feedback schedules~(full 6800-sample schedule, and 1/2, 1/4 and 1/6 thereof) and learning setups~(no demonstrations, demonstrations + preferences, demonstrations + preferences + initial automatic labels from demonstrations, and demonstrations with non-synthetic, actual human preferences). This information is synthesized in \autoref{fig:effort_time_max}, where the best achievable performance for a given amount of effort is displayed, either in the no-demonstration setup or in setups that make use of demonstrations.

The more demo-driven games~(Hero, Montezuma's Revenge, Private Eye and Q*bert) are also more feedback-driven, improving with additional feedback if demonstrations are available. The opposite is true for Enduro, where the score increases with additional feedback only if demonstrations are not used. The agent easily beats the demonstrations in this game, so they work against the feedback. Pong is solved with very little synthetic feedback in any setup, but when preferences are provided by a human it significantly improves with extra feedback. Beamrider without demonstrations peaks at low feedback and then regresses, but with demonstrations it is feedback-driven. Breakout and Seaquest don't display a clear trend. Breakout is especially noisy~(see variance in \autoref{fig:all_bars}), the score depending greatly on whether the agent discovers `tunneling behind the wall', absent in the demonstrations. In Seaquest the agent does not learn to resurface, limiting the scores in all setups. The reason for this is the removal of episode boundaries: the small penalty we add at the end of a game is offset by the extra time the agent can spend shooting enemies if it does not resurface. Furthermore, preferences fail to accurately capture the difference in reward magnitude between collecting all divers and bringing them to the surface compared to shooting enemies.

\section{Reward model training}
\label{sec:predictor_loss}

\begin{figure}[h!]
\begin{center}
\includegraphics[scale=0.2]{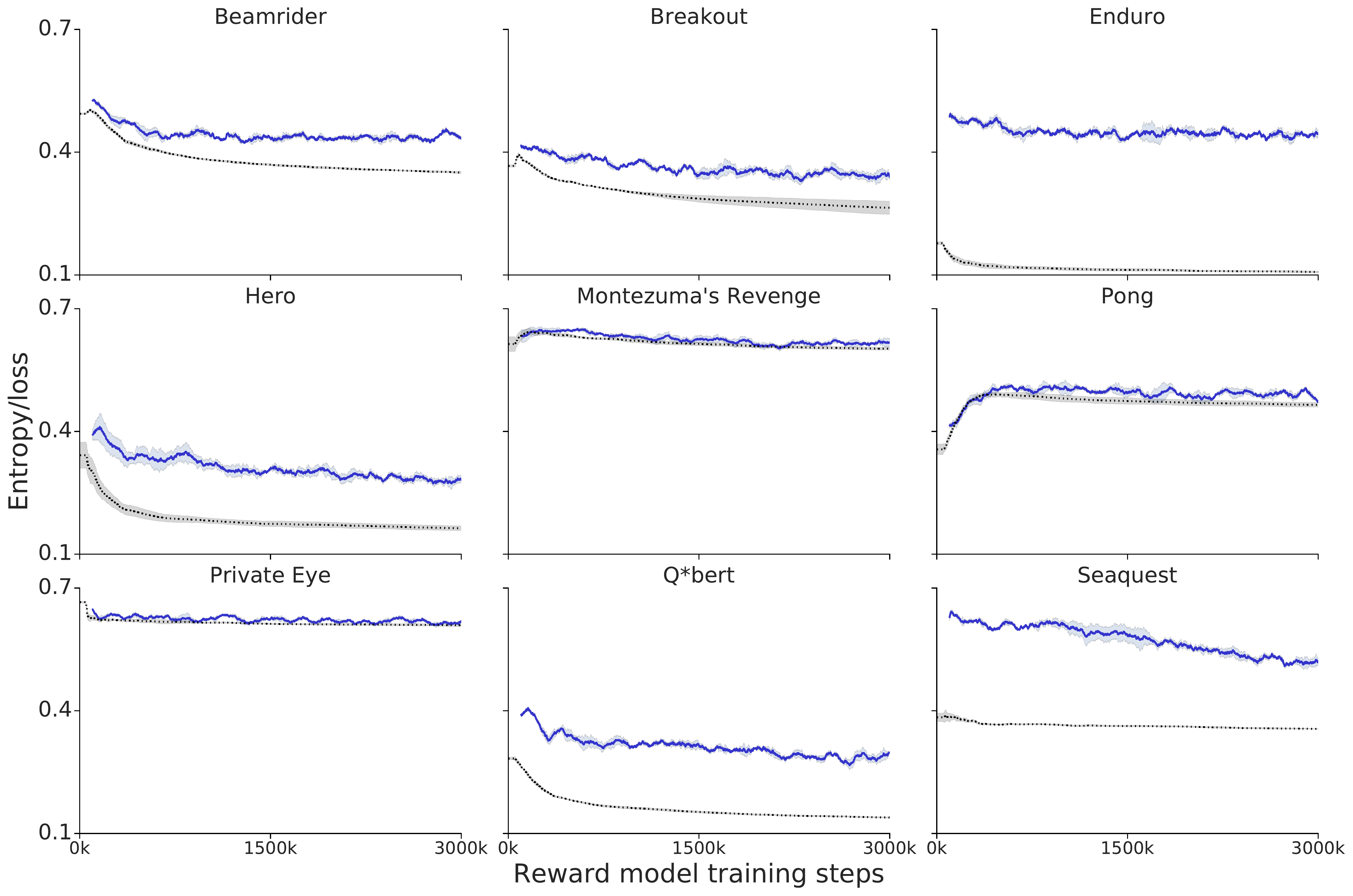}
\end{center}
\caption{Cross-entropy loss of the reward model~(solid blue line) and average label entropy~(dotted black line) during training. The average label entropy is a lower bound for the loss and depends on the ratio of `indifferent' labels in the annotated comparisons: more `indifferent' labels result in higher entropy.}
\label{fig:predictor_loss}
\end{figure}

\autoref{fig:predictor_loss} depicts the reward model training loss, which is the cross-entropy between the two-class labels from clip comparison and the model output. The labels can be either (1, 0) when the first clip is preferred, (0, 1) when the second clip is preferred, or (0.5, 0.5) when neither clip is preferred (`indifferent' label). `Indifferent' labels have minimum possible cross-entropy $\log\ 2=0.693$, so the loss has a lower bound that depends solely on the ratio of `indifferent' labels in the annotation dataset. This ratio varies significantly from game to game. Games with sparse rewards, like Private Eye and Montezuma's Revenge, have a high proportion of indifferent clip pairs (both segments lacking any reward) and therefore high lower bounds. The bound evolves during training as more labels are collected.

We expect a well-trained reward model to stay close to the label entropy bound. This is to some extent what happens with Beamrider, Breakout, Montezuma's Revenge and Private Eye. In other games like Hero, Q*bert and Seaquest the loss is between $50$\% and $100$\% above the entropy bound. The game with seemingly worst reward model training is Enduro, where the loss is more than $4$ times the lower bound. This big gap, however, can be explained by the fine-grained scoring of the game. Points are earned continuously at a rate proportional to the car speed, so very similar clips can differ by one point and not be labeled `indifferent' by the synthetic annotator. The reward model does not latch to these preferences for similar clips, but learns to distill the important events, such as crashes that significantly reduce the accumulated score. As a result, the predicted reward is highly correlated with the game score and the agent learns to play the game well, even if the model fails to tease apart many pairs of clips with small score differences.

\section{Reward model alignment}
\label{sec:reward_alignment}

\begin{figure}[h!]
\begin{center}
\includegraphics[scale=0.2]{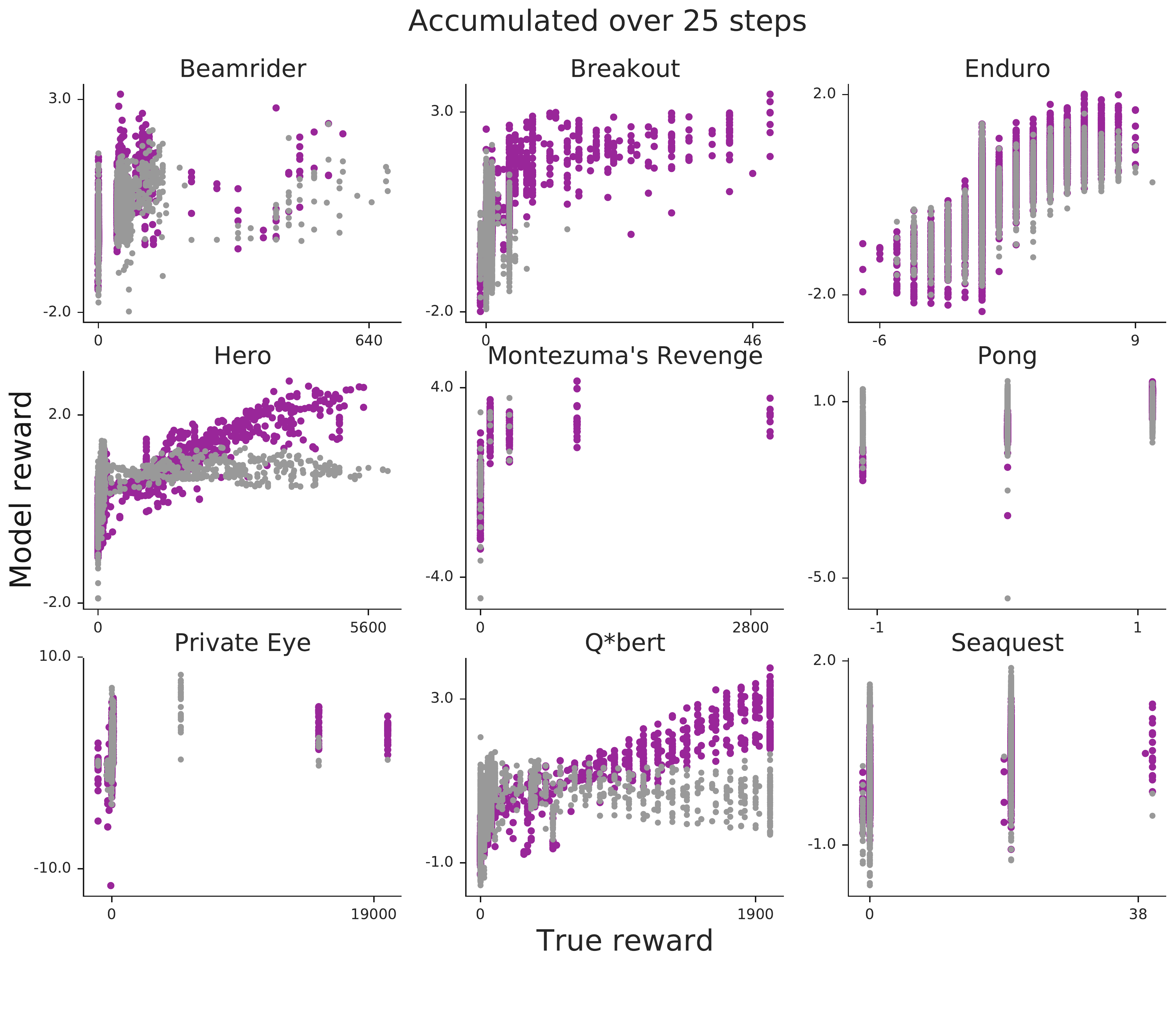}
\includegraphics[scale=0.2]{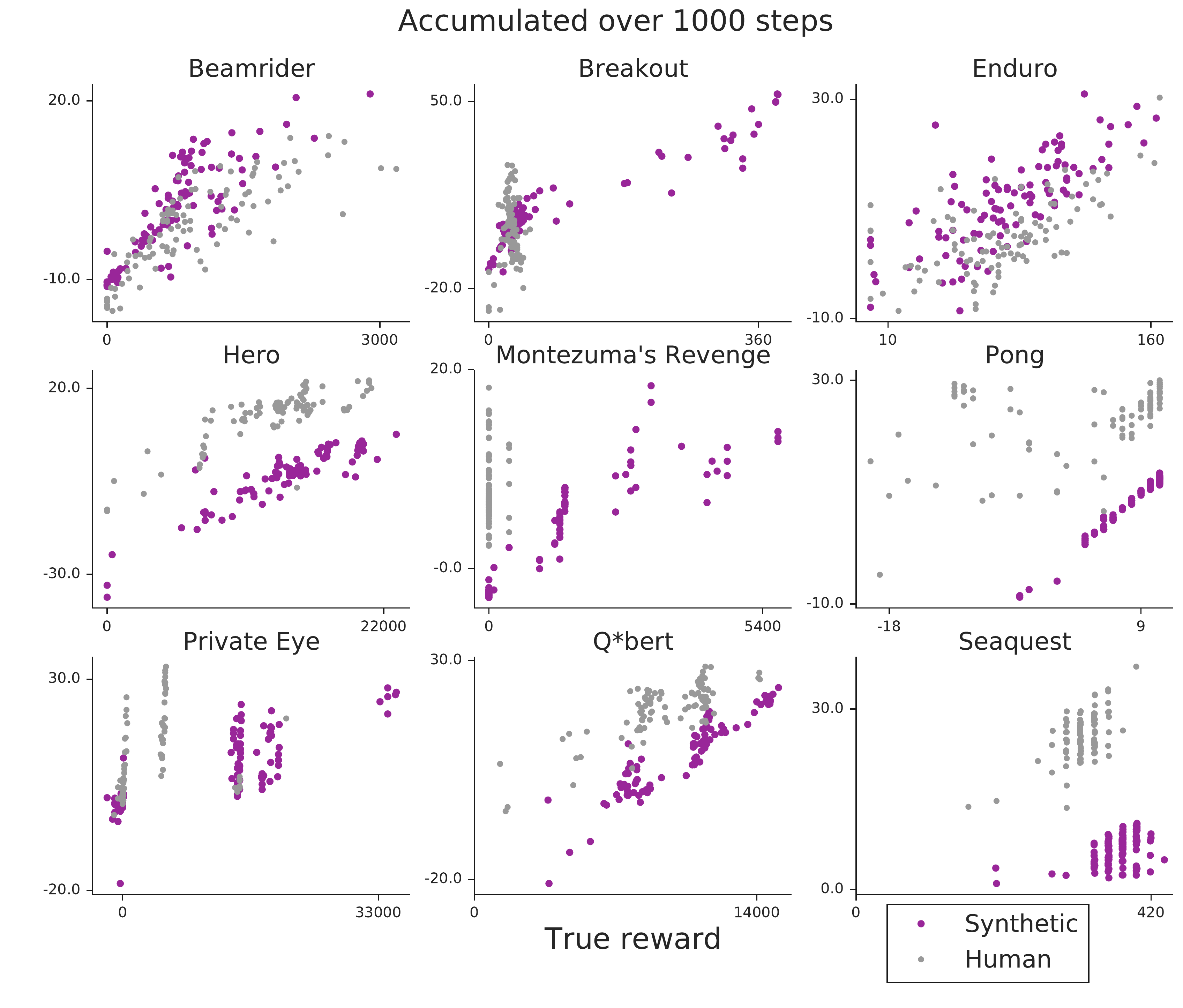}
\end{center}
\caption{True vs.\ model reward accumulated in sequences of $25$~(left) and $1000$~(right) agent steps. Magenta and gray dots represent the model learned from synthetic~(demos + pr.\ + autolabels in \autoref{fig:all_bars}) and human preferences, respectively. A fully aligned reward model would have all points on a straight line. For this evaluation, the agent policy and reward model were fixed after successful full-schedule training (in the case of synthetic feedback we chose the most successful seed; in the case of human feedback, there was no choice, only one run was available).}
\label{fig:reward_alignment_2}
\end{figure}

\autoref{fig:reward_alignment_2} displays model reward plotted against true reward, accumulated over short~($25$ agent steps) and long~($1000$ agent steps) time intervals, for experiments with synthetic preference feedback and human preference feedback. With synthetic preferences, the reward model generally aligns well with the true reward, especially over longer time intervals. We can observe the following:
\begin{itemize}
\item Sparse rewards (as in Montezuma's Revenge and Private Eye) make preference learning harder because they result in fewer informative preference labels.
\item Learning different reward sizes with preferences is hard because preferences do not express precise numeric feedback. A reward of 10 vs. 11 generates the same label as 10 vs. 100; to learn the difference the model needs to chance upon pairs of clips linked by the intermediate reward.
\item The coarser alignment over short intervals makes the learned model hackable in games where imprecisely timed rewards can be exploited (see~\autoref{fig:fixed_predictor}).
\end{itemize}

As for human preferences, the reward model fails to align with the true reward in Breakout, Montezuma's Revenge and Pong. One reason is that the reward function the human wants to teach can be different from the Atari simulator's. For example, in Montezuma's Revenge the human tries to shape the reward by punishing deaths~(which are not punished according to the true reward), resulting in a passive policy. In Pong and Breakout, the human gives priority to hitting the ball rather than scoring points, so sometimes it is worth dropping the ball to score an easy hit on restart.

\section{Effects of label noise}
\label{sec:label_noise}

\begin{figure}[h!]
\begin{center}
\includegraphics[scale=0.25]{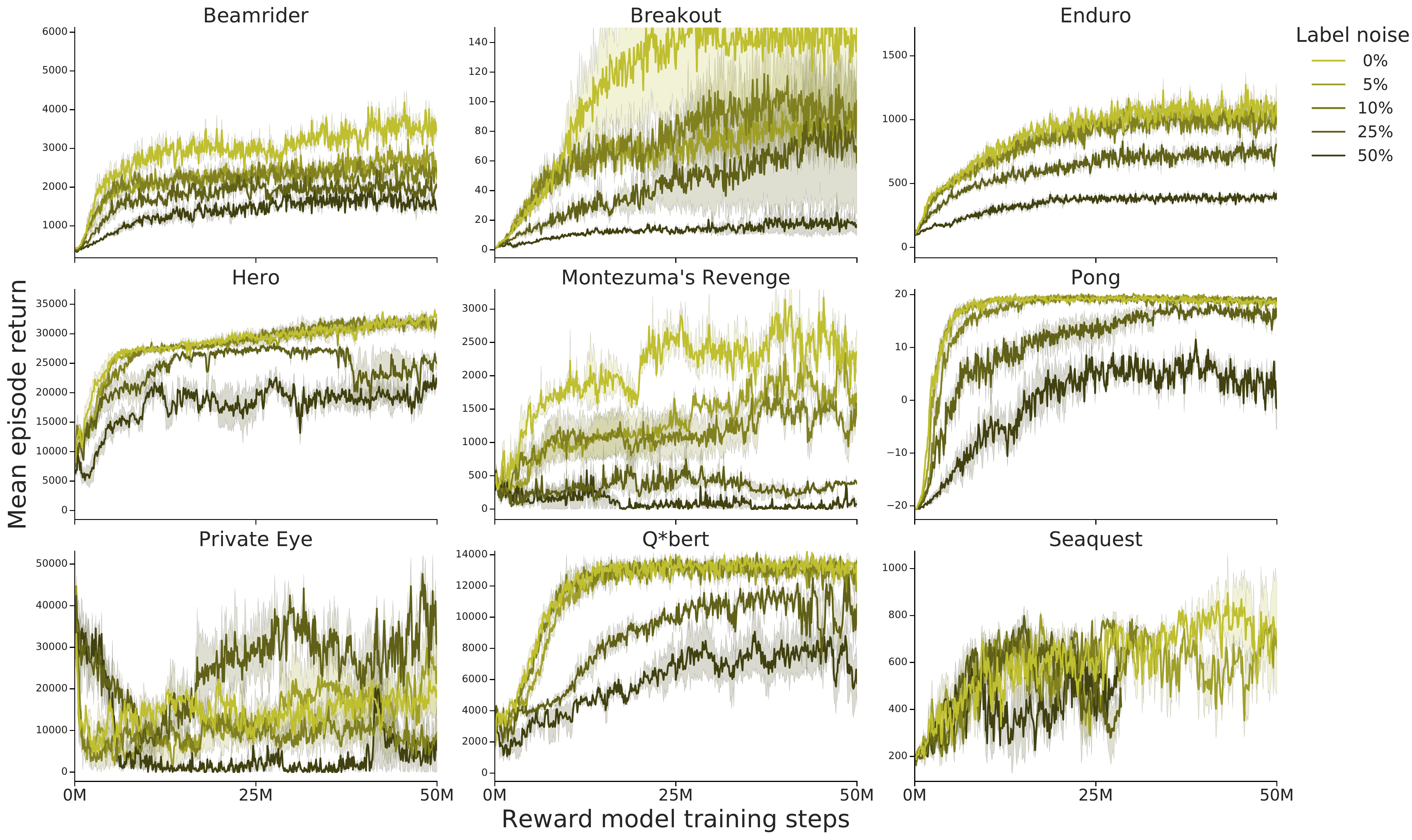}
\end{center}
\caption{Performance in each game with different rates of mislabelling by the annotator. Experiments are from synthetic feedback with full label schedule and not using automatic labels from demonstrations~(same setting as `human' experiments).}
\label{fig:label_noise}
\end{figure}

The relatively poor performance of human compared to synthetic annotators can be partially explained by random mistakes in labeling. \autoref{fig:label_noise} shows
how the different games are affected by labeling mistakes. The mistake rates observed in our human-annotated experiments are between $5\%$ and $10\%$. Those noise levels have minor impact in most games, but they are significantly detrimental in Montezuma's Revenge, and thus partially accounts for the poor results in the human-labeled experiments.

\section{Comparison with policy gradient}
\label{sec:compare_a3c}

\citet{Christiano17} used the policy-gradient-based agent A3C to evaluate preference feedback~(without demonstrations). In this paper we used the value-based agent DQN/DQfD. The following table compares our scores without demonstrations with corresponding scores in \citet{Christiano17}.

\begin{center}
\begin{tabular}{p{0.13\textwidth}|c|c||c|c|}
 & \multicolumn{2}{c||}{\textbf{DQN + preferences~(ours)}} & \multicolumn{2}{c|}{\textbf{A3C + preferences}}\\
Game & 6800 labels & 3400 labels & 5500 labels & 3300 labels\\
\hline
Beamrider & 3000 & 4000 & \textbf{10000} & \textbf{10000} \\
Breakout & \textbf{100} & 40 & 20 & 20 \\
Enduro & \textbf{1600} & 1400 & 0 & 0 \\
Pong & 19 & 19 & \textbf{20} & \textbf{20} \\
Q*bert & 5800 & 7800 & \textbf{13000} & 5000 \\
Seaquest & 1000 & 800 & \textbf{1200} & 800 \\
\end{tabular}
\end{center}

\section{Comparison with DQfD trained from true reward}
\label{sec:compare_dqfd}

The following table compares the average scores of DQfD~\citep{Hester18} trained from true reward and from a learned reward model~(ours). Our scores are from full schedule runs with autolabels~(magenta bars in \autoref{fig:all_bars}).

\begin{center}
\begin{tabular}{l|c|c||l|c|c|}
Game & \multicolumn{1}{|p{0.15\textwidth}|}{\textbf{DQfD + feedback~(ours)}} & \multicolumn{1}{|p{0.13\textwidth}||}{\textbf{DQfD + true reward}} &
Game & \multicolumn{1}{|p{0.15\textwidth}|}{\textbf{DQfD + feedback~(ours)}} & \multicolumn{1}{|p{0.13\textwidth}|}{\textbf{DQfD + true reward}} \\
\hline
Beamrider & 4100 & \textbf{5170} & Pong & \textbf{19} & 11 \\
Breakout & 85 & \textbf{310} & Private Eye & \textbf{52000} & 42500 \\
Enduro & 1200 & \textbf{1930} & Q*bert & 14000 & \textbf{21800} \\
Hero & 35000 & \textbf{106000} & Seaquest & 500 & \textbf{12400} \\
Montezuma's & 3000 & \textbf{4640}
\end{tabular}
\end{center}

Note that this does not compare like with like:
while training with a synthetic oracle makes the true reward function indirectly available to the agent, our method only uses a very limited number of preference labels~(feedback on at most $340.000$~agent steps), providing reward feedback on $<1\%$ of the agent's experience. To make a fair comparison with DQfD, we should only allow DQfD to see the reward in $1\%$ of the training steps. This would result in very poor performance~(results not reported here).

\section{Unsuccessful ideas}
\label{sec:unsuccessful-ideas}

In addition to the experiments presented in the main paper, we were unsuccessful at getting improvements from a variety of other ideas:
\begin{enumerate}
\item Deep RL is very sensitive to the scale of the reward, and this should be alleviated with quantile distributional RL while improving performance~\citep{Dabney17}. However, we did not manage to stabilize the training process with either distributional RL~\citep{Bellemare17} or quantile distributional RL.
\item Both reward model and policy need to learn vision from scratch and presumably share a lot of high-level representations. Previous work has shown that training the same high-level representations on multiple objectives can help performance~\citep{Jaderberg16}. However, weight sharing between policy and reward model as well as copying of weights from the policy to the reward model destabilized training in our experiments.
\item To improve the reward models sample efficiency, we used parts of a pretrained CIFAR convolutional network as well as randomly initialized convolutional network. While this provided slight improvements in sample-efficiency on a few games, the effect was not very pronounced.
\item Since every observation from the environment is an unlabeled data point for the reward model, we could leverage techniques from semi-supervised learning to improve reward model's sample complexity. We tried applying the state of the art technique by \citet{Tarvainen17} without much improvement in sample complexity while facing more training stability issues. Unfortunately it is unclear whether that particular approach does not work very well on the Atari visuals or whether the problem structure of reward learning is not very amenable to semi-supervision~(for example because the reward is not very continuous in the visual features).
\item The bias toward expert demonstrations can limit the agent performance in environments where the expert performs poorly. We ran experiments where the large-margin supervised loss $J_E$ was gradually phased out during training. This had the desired effect of boosting performance in Enduro, where demonstrations are detrimental, but in games where demonstrations are critical, like Montezuma's Revenge and Private Eye, performance dropped along with the phasing-out of the supervised loss.
\item When using the expert demonstrations to augment preference annotation, we tried requesting annotations on pairs made up of two demo clips as a way to cover more of the state space in the reward model. It did not change the performance.
\item We also accidentally noticed that DQfD is extremely sensitive to small differences between the demonstration observations and the agent's experience---a small misalignment in the score-blanking method between demos and policy frames reduced the agent's scores to zero in the demo-heavy games. We attempted to increase robustness by adding noise and more regularization to the agent, but all such attempts hurt performance significantly.
\end{enumerate}

\end{document}